\documentclass[11pt]{article}

\usepackage[preprint]{acl}

\usepackage{times}
\usepackage{latexsym}

\usepackage[T1]{fontenc}

\usepackage[utf8]{inputenc}

\usepackage{microtype}
\usepackage{subcaption}

\usepackage{inconsolata}
\usepackage{amsmath}
\usepackage{amssymb}
\usepackage{booktabs}

\usepackage{graphicx}
\usepackage{multirow}

\title{A Unified Response Framework for Questions and Tutoring Dialogues}

\author{
  \textbf{Zhangqi Duan}\textsuperscript{1},
  \textbf{Simon Woodhead}\textsuperscript{2},
  \textbf{Maya Miller-Vedam}\textsuperscript{3},
  \textbf{Andrew Lan}\textsuperscript{1} \\
  \textsuperscript{1}University of Massachusetts Amherst,
  \textsuperscript{2}Eedi,
  \textsuperscript{3}Khan Academy\\
  \texttt{\{zduan, andrewlan\}@cs.umass.edu},\\
  \texttt{simon.woodhead@eedi.co.uk},\\
  \texttt{mayavedam@khanacademy.org}
}

\begin{document}
\maketitle
\begin{abstract} 
Existing student models in educational applications largely focus on student responses to questions. However, as tutoring becomes more accessible at large scale, it is important to analyze dialogues between tutors and students, which may reveal fine-grained evidence on student knowledge. In this work, we propose a unified framework that models both student responses to questions and tutor-posed subtasks during dialogues on a shared item response theory (IRT) scale. We first identify mathematical subtasks in tutoring dialogues, rewrite them into self-contained questions, and identify the correctness of student responses to these subtasks. We then develop an IRT framework that estimates question/subtask difficulty from text, tracks students' evolving abilities across question-answering and dialogue interactions, and a novel method to leverages tutor moves between consecutive subtasks to regularize their difficulty values. We evaluate our framework on two real-world math education datasets containing both human- and AI-tutored dialogues. Results show that our framework outperforms strong baselines on both question difficulty prediction and response correctness prediction. 

\end{abstract}

\section{Introduction}

Estimating student knowledge and item characteristics from students' responses to mathematical questions/items is central to online learning platforms, which supports personalized practice, targeted tutoring, and reliable assessment. There are two major types of student response modeling approaches: Item response theory (IRT) models place static student ability and question difficulty on a common scale \cite{rasch1960probabilistic}, while knowledge tracing (KT) models track how a student's knowledge evolves over time across a sequence of questions \cite{corbett1994knowledge}. One key limitation of these models is that they only account for students' response records, which may leave out fine-grained information on their knowledge. For example, for multiple-choice questions, only knowing whether a student selected the correct option or which incorrect option they selected does not allow us to pinpoint their misunderstanding. 

\begin{figure}[t!]
    \centering
    \includegraphics[width=1\linewidth]{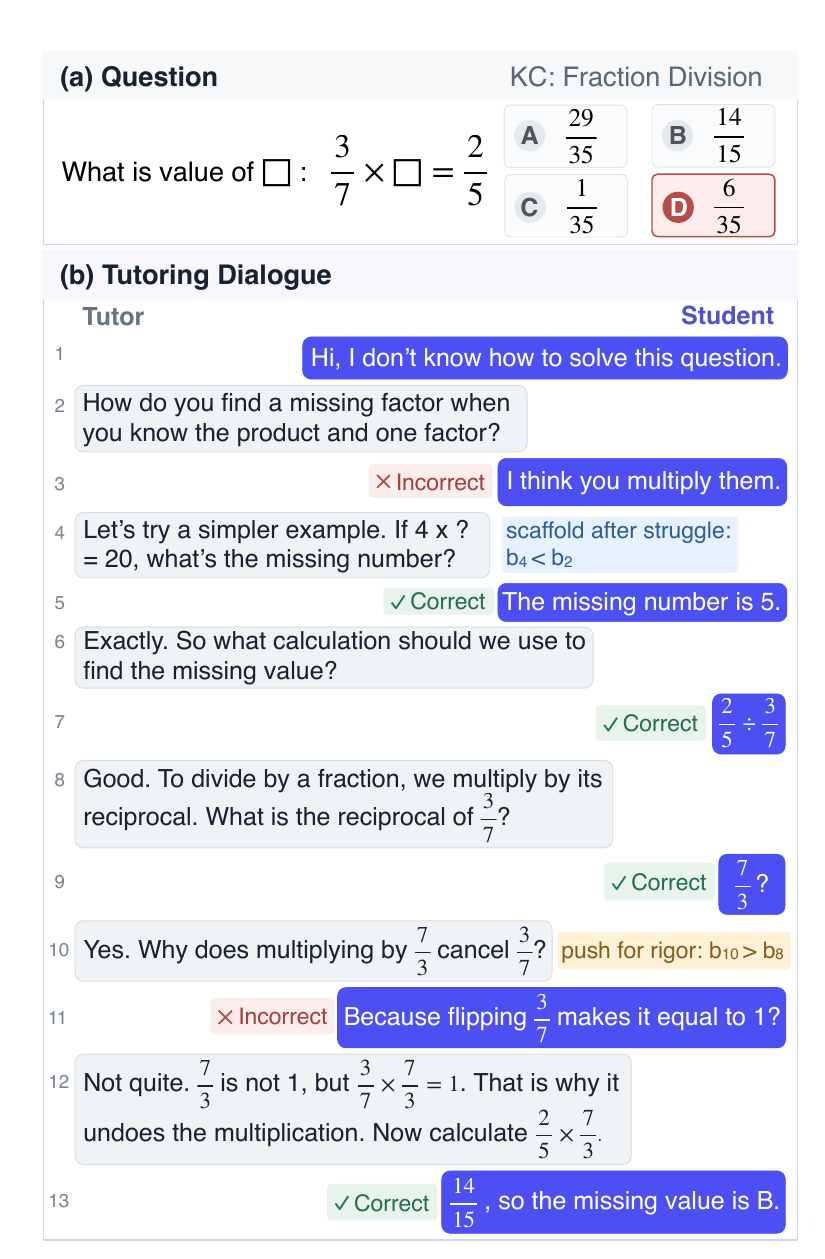}
    \caption{Student responses to tutor-posed subtasks in tutoring dialogues reveal fine-grained insights into the student's changing knowledge, while tutor moves help infer the relative difficulty of subtasks.}
    \label{fig:motivation}
    \vspace{-.5cm}
\end{figure}

Powered by generative artificial intelligence (AI), online learning platforms increasingly combine standalone question-answering practice with on-demand tutoring \cite{eedi,khanmigo,livehint}. Tutoring dialogues have significant potential in providing us fine-grained insights into student knowledge: a tutoring dialogue is not merely an explanation attached to a question; it is a sequence of smaller assessment and learning events. A tutor may ask the student to recall a fact, perform an intermediate calculation, or justify a step; after observing the response, the tutor may scaffold or push the student toward more rigorous reasoning \cite{tutormoments}. The student's knowledge can therefore be more accurately estimated and may change throughout the dialogue. Figure~\ref{fig:motivation} illustrates this process. The original response only indicates that the student answered the fraction-division question incorrectly, whereas the dialogue reveals an initial conceptual error, success on a simpler scaffolded task, correct intermediate calculations, and a remaining gap in justification. These \emph{tutor-posed} tasks and student responses may provide substantially more insights into the student's latent knowledge than the original response alone.

\paragraph{Related Work}
However, existing approaches do not provide a rigorous way to analyze original question responses and tutor-posed tasks together. Prior work has explored question difficulty prediction from question content~\cite{huang2017question,benedetto2021transformers} and simulated student responses~\cite{scarlatos-etal-2025-smart}, as well as KT in tutoring dialogues~\cite{Scarlatos_2025}. More recently,~\citet{huang2026interpretabledifficultyawareknowledgetracing} incorporate tutor-turn difficulty and IRT into conversational KT. However, two challenges remain for jointly modeling original questions and dialogue subtasks. First, tutoring dialogues follow personalized trajectories, resulting in subtasks that are often not shared across students and thus difficult to calibrate using traditional IRT. Second, many tutor-posed tasks implicitly refer to quantities, expressions, or reasoning steps introduced earlier in the problem or dialogue, making them difficult to represent as self-contained, standalone questions. Together, these challenges leave open the question of how to place original questions and tutor-posed tasks on the same scale and use them to track student knowledge. We provide a detailed discussion of related work in Appendix~\ref{appendix:related_work}.

\paragraph{Contributions}
In this work, we propose \textbf{URF-QD} (Unified Response Framework for Questions and tutoring Dialogues), a framework for jointly modeling responses to original questions and mathematical subtasks arising in corresponding tutoring dialogues. Our contributions are threefold. First, we identify tutor turns that pose mathematical tasks and rewrite these context-dependent tasks as concise, self-contained questions, enabling their difficulties and student responses to be modeled jointly with those of original questions. Second, we introduce a dynamic IRT model that tracks the student's evolving knowledge state from responses to both original questions and dialogue subtasks; importantly, we introduce a novel tutor move-implied regularization term to supervise a relative ordering between consecutive subtasks. Third, we evaluate our framework on two real-world math education datasets containing both question-answering records and tutoring dialogues. Extensive evaluation shows that our framework consistently improves question difficulty prediction and response correctness prediction. We use human evaluation and qualitative analyses to validate and illustrate components of our framework, and present a preliminary example of how our framework can be used for large language model (LLM) tutor benchmarking in the Appendix. Our implementation will be made publicly available to enable future research on student assessment from dialogues. 

\begin{figure*}
    \centering
    \includegraphics[width=1\linewidth]{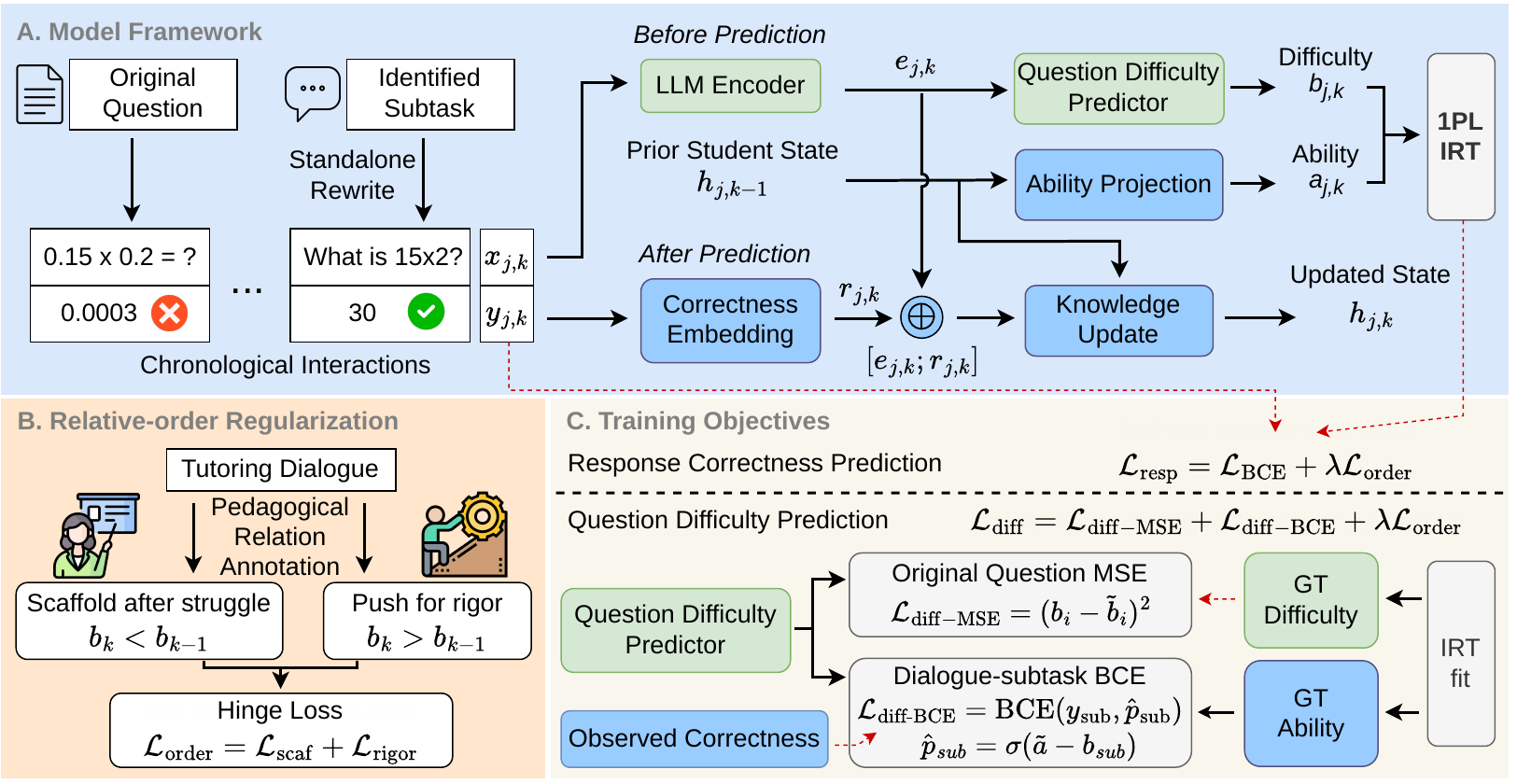}
    \caption{Overview of URF-QD. (A) Original questions and rewritten dialogue subtasks are modeled on a shared IRT scale using text-based difficulty estimates and evolving student ability. (B) Relative-order regularization encourages scaffolded tasks to be easier and rigor-promoting tasks harder than preceding tasks. (C) Response correctness uses BCE with the regularization loss, while question difficulty uses GT-difficulty MSE, dialogue-subtask BCE, and the regularization loss. GT denotes IRT-estimated “ground-truth” question difficulties and student abilities obtained from observed student responses to original questions.}
    \label{fig:main_fig}
    \vspace{-.5cm}
\end{figure*}

\section{Problem Formulation}
We consider the setting of students working in digital learning platforms, where they practice on math questions and interact with tutors when they want additional support. Our goal is to estimate a student's ability from how they respond to both \textbf{questions} and \textbf{tutor-posed subtasks} that arise naturally during tutoring dialogues.

For each student $j$, we denote the binary-valued correctness of their response to an original question $q_i$ as $y_{i,j} \in \{0, 1\}$, where $1$ corresponds to a correct response. When receiving tutoring support on question $q_i$, the student and the tutor engage in a dialogue $d_{i, j}=(s_0, t_1, s_1, ... , t_M, s_M)$, where $s_m$ and $t_m$ denote student and tutor utterances respectively and $M$ is the number of total tutor-student turn pairs. The initial student turn $s_0$ is present when the student initiates the conversation.

We focus on tutor turns that pose a mathematical subtask, i.e., when the tutor requires the student to produce a mathematically meaningful response, such as an intermediate calculation, factual recall, method selection, interpretation, or explanation. We treat student responses to these tutor-posed tasks as additional response observations. We detail how these tasks are identified and transformed into standalone questions in Section~\ref{subtask_rewrite}. 

For each student $j$, we order all scored interactions chronologically and denote the resulting sequence as $H_j=((x_{j,1}, y_{j,1}), ..., (x_{j,T}, y_{j,T}))$, where $x_{j,k}$ denotes the mathematical task associated with interaction $k$, and $y_{j,k} \in \{0,1\}$ denotes the corresponding response correctness. An interaction may correspond either to an original question or to a subtask arising within a tutoring dialogue. 

Given the interaction history $H_{j, <k}$, our primary modeling objective is to estimate $p(y_{j, k}=1|x_{j,k}, H_{j, <k})$, while placing student ability and the difficulties of original questions and dialogue tasks on a common IRT scale. In this way, question-answering and tutoring-dialogue interactions provide complementary observations for modeling student performance. 

\section{Methodology}
\label{sec: methodology}
We now detail our framework for jointly modeling student performance in question-answering activities and tutoring dialogues. Our key idea is to place original questions and tutor-posed dialogue subtasks on the same IRT parameter scale. As shown in Figure~\ref{fig:main_fig}, our framework has three main components. First, we identify mathematical subtasks posed by tutors during dialogues and transform these context-dependent tasks into the format of standalone questions. Second, we use a dynamic IRT model to jointly model responses to original questions and transformed dialogue subtasks on a shared scale. Third, we leverage tutor pedagogical move labels to impose difficulty ordering constraints between consecutive tutor-posed tasks, to provide additional supervision when estimating their difficulties. We note that, for simplicity, we restrict ourselves to 1PL IRT models \cite{hambleton2013item}, which is often referred to as the Rasch model \cite{rasch1960probabilistic}, where each question is only characterized by a single scalar-valued difficulty parameter. Our framework can be extended to other IRT models, although difficulty estimation is known to be more accurate than scale/discrimination estimation for 2PL IRT models \cite{chen2026from}.

\subsection{Subtask Transformation}
Tutoring dialogues contain many intermediate mathematical tasks that provide additional evidence of student ability. However, tutor utterances are often context-dependent and differ substantially in form from standalone assessment questions. We therefore first identify tutor turns that constitute mathematical subtasks and then rewrite each identified subtask into a standalone question, allowing original questions and subtasks to be represented in a common form for subsequent modeling.

\paragraph{Subtask Identification} For each dialogue $d_{i,j}$, we consider every tutor turn that is immediately followed by a student turn as a candidate subtask. We prompt GPT-5.4 with the complete dialogue to identify the valid mathematical subtasks: for each candidate, it determines whether the tutor turn poses an explicit mathematical task and whether the subsequent student turn attempts to answer it. We retain only candidates that satisfy both conditions. Following \cite{scarlatos-etal-2026-simulated}, for each retained tutor subtask, we annotate whether the student response $s_m$ correctly answers the tutor-posed task. For each scored response, we denote this correctness label as $y_{i,j,m} \in \{0,1\}$. Turns without an applicable mathematical response are left unscored and are not used as response observations.

\paragraph{Standalone Question Transformation} 
\label{subtask_rewrite}
For every identified tutor subtask \(t_m\), we prompt GPT-5.4 to rewrite the tutor's request as a concise, standalone assessment question \(u_{i,j,m}\). Given the original problem and dialogue with the target turn marked, the model resolves contextual references and supplies only the information needed to make the question self-contained while preserving its mathematical demand. We provide further prompt details in Appendix~\ref{sec: prompts_appendix}.

For each scored subtask, we map the transformed question \(u_{i,j,m}\) and its response label $y_{i,j,m}$ to the corresponding interaction $k$ in the student's chronological history $H_j: x_{j,k}=u_{i,j,m}, y_{j,k}=y_{i,j,m}.$ This mapping represents dialogue subtasks in the same form as original questions, enabling both interaction types to be modeled jointly. 

\subsection{URF-QD: Joint Modeling of Questions and Dialogue Subtasks}
Given the chronological interaction history $H_j$, which contains both original question responses and transformed dialogue subtask interactions, we model student performance using a dynamic IRT framework. We process interactions sequentially and maintain a latent student state $\mathbf{h}_{j,k}$ that summarizes the student's interactions up to step $k$. At interaction $k$, we first encode the current task $x_{j,k}$ and estimate its difficulty. We then project the preceding state $\mathbf{h}_{j,k-1}$ to a scalar ability and use a 1PL IRT model to predict the student's response. After observing $y_{j,k}$, we update the student state using the task representation and response correctness.

\paragraph{Task Encoder and Difficulty} 
We encode each task $x_{j,k}$ using an LLM. Let
$\mathbf{Z}_{j,k}=(\mathbf{z}_{j,k,1},\ldots,\mathbf{z}_{j,k,L_{j,k}})$
denote the last-layer hidden state representations of the question sequences $L_{j,k}$. We obtain a fixed-dimensional task representation by mean pooling: $
\mathbf{e}_{j,k}
=
\frac{1}{L_{j,k}}
\sum_{\ell=1}^{L_{j,k}}
\mathbf{z}_{j,k,\ell}.
$
We map this representation to a scalar difficulty using a multilayer perceptron (MLP) $g_{\mathrm{diff}}$:
$
b_{j,k}=g_{\mathrm{diff}}(\mathbf{e}_{j,k}).
$
The task encoder and difficulty prediction head are shared across original questions and dialogue subtasks, placing both on the same latent difficulty scale. This text-based parameterization is particularly important for dialogue subtasks, since different students may encounter different tutor-posed tasks even when working on the same original question. Rather than assigning an independent difficulty parameter to every observed task, the model estimates difficulty from its textual content, enabling the learned difficulty function to generalize across questions, students, and dialogues.

\paragraph{Dynamic Student State and Ability}
We represent student $j$'s interaction history with a latent state $\mathbf{h}_{j,k}$ that summarizes all observed task-response pairs through step $k$. After observing the student's response $y_{j,k}$ to task $x_{j,k}$, we incorporate this new evidence into the state. We first represent response correctness using a learned embedding:
$\mathbf{r}_{j,k}=E_y(y_{j,k}),$ where $E_y$ contains separate learnable embeddings for correct and incorrect responses to both original questions and dialogue subtasks. We then concatenate resulting correctness embedding with task's textual embedding. One can use any deep learning-based KT method for this dynamic update; unless otherwise noted, we defer to the deep KT (DKT) method \cite{piech2015deepknowledgetracing}. 

\paragraph{IRT Response Model}
Before the student attempts task $x_{j,k}$, we estimate their current ability by projecting the preceding state to a scalar: $a_{j,k}=g_a(\mathbf{h}_{j,k-1}),$
where $g_a$ is a MLP. This pre-interaction ability is used in the IRT response model below and does not depend on the current response $y_{j,k}$. After that response is observed, the updated state $\mathbf{h}_{j,k}$ is used to estimate ability at interaction $k+1$. Unlike a static IRT ability parameter, $a_{j,k}$ is dynamic throughout the student's interaction sequence and updated when new responses are observed. Therefore, dialogue subtask responses serve not only as correctness-prediction targets, but also as additional observations that refine future ability estimates.
Given the student's current ability $a_{j,k}$ and the difficulty of the current task, $b_{j,k}$, we model response correctness using a 1PL logistic IRT model:
\[
p(y_{j,k}=1 \mid x_{j,k}, H_{j,<k})
=
\sigma(a_{j,k}-b_{j,k}),
\]
where $\sigma(\cdot)$ denotes the sigmoid function. 
Because original questions and dialogue subtasks share the same task encoder, difficulty head, ability projection, and IRT response function, their difficulties and the student's dynamic ability are represented on a common latent scale.

\subsection{Difficulty Order Regularization}
Correctness observations supervise the relationship between student ability and task difficulty, but tutoring dialogues provide an additional source of information: the tutor's pedagogical move can imply a relative difficulty ordering between consecutive subtasks. We exploit two cases, \textbf{scaffolding after struggle} and \textbf{pushing for rigor} \cite{tutormoments}, and use them to regularize task difficulties.

\paragraph{Pedagogical Move Annotation} For each pair of consecutive tutor subtasks separated by a single student response, we prompt GPT-5.4 to determine whether the transition implies a tutor move that implies any relationship between the subtask difficulties. The model is provided with the original problem, the raw tutoring dialogue with subtasks marked, eligible tutor-turn pairs, and the annotated correctness label of the intervening student response. We withhold the student's response to the second subtask, so that the relation is determined from the tutor's pedagogical action rather than from whether the student subsequently succeeds or fails.

We denote the set of turns that correspond to scaffolding after struggle moves as \(\Omega_1\). In practice, when students respond incorrectly to tutor-posed subtasks or indicate that they do not know the answer, tutors often scaffold by breaking down the original subtask into smaller, simpler, more concrete subtasks; such moves imply that $b_{j,k} < b_{j,k-1}$. Similarly, we denote the set of pushing for rigor turns as \(\Omega_2\). A pair is assigned to \(\Omega_2\) when the previous student response is correct and the subsequent tutor subtask increases the required depth of reasoning, for example by eliciting a justification, explanation, derivation, interpretation, proof, or other conceptual reasoning; such moves imply $b_{j,k} > b_{j,k-1}$. 
We assign no constraint when the intervening response cannot be scored for correctness, when the two subtasks are unrelated, or when the second tutor turn merely repeats the task, requests a generic retry, continues a routine multi-step solution, or otherwise does not substantively change the cognitive demand. 

\paragraph{Regularization} For each \((j,k)\in\Omega_1\), the second task should be easier than the preceding task. We define the corresponding hinge loss as \[
\mathcal{L}_{\text{scaf}} = \frac{1}{|\Omega_1|}\sum_{(j,k) \in \Omega_1} \max\{b_{j,k}-b_{j,k-1}, 0\}.
\]
The loss is positive when the scaffolded task is more difficult than the previous task, violating the tutor move, and zero otherwise. We define the loss for pushing for rigor moves, $\mathcal{L}_{\text{rigor}}$, in a similar way, with $b_{j,k}$ and $b_{j,k-1}$ reversed. 
%
%
We simply sum these two regularization terms together to obtain $\mathcal{L}_{\mathrm{order}}$ and add it to the objectives detailed below, using them as soft penalties to encourage the difficulty encoder to learn dialogue subtask difficulties that are consistent with tutoring moves. 

\subsection{Prediction Targets, Training Objectives}
We study two complementary prediction targets using separate training objectives. \emph{Question difficulty prediction} evaluates whether the shared difficulty encoder can accurately estimate the difficulty of previously unseen questions. \emph{Response correctness prediction} evaluates whether original-question and dialogue-subtask responses can jointly track a student’s evolving ability and accurately predict their future performance. 

\paragraph{Question Difficulty Prediction}
Here, we examine whether dialogue subtasks provide additional calibration to help us estimate the difficulty of original questions. We first fit an IRT model on observed student responses and use it to obtain a ``ground-truth'' difficulty $\tilde{b}_i$ for each original question $q_i$. 
We then train the shared difficulty predictor to recover these difficulties, while using dialogue interactions as additional supervision.
For an original question $q_i$, let $b_i$
denote the difficulty value predicted by the encoder given the question content as input. We supervise these predictions using mean squared error over the original questions available for training, as $
\mathcal{L}_{\mathrm{diff\text{-}MSE}}
=
\frac{1}{|\mathcal{Q}_{\mathrm{train}}|}
\sum_{i\in\mathcal{Q}_{\mathrm{train}}}
\left(b_i-\tilde{b}_i\right)^2.
$

Since dialogue subtasks do not have corresponding ground-truth difficulty labels, we additionally train on the observed student response correctness to extract information on subtask difficulty indirectly. Let $\tilde{a}_j$ denote the fixed reference ability of student $j$ from the fitted IRT, and let $b_{i,j,m}$
be the predicted difficulty of transformed dialogue subtask $u_{i,j,m}$, we minimize $\mathcal{L}_{\mathrm{diff\text{-}BCE}}$, the standard binary cross-entropy (BCE) loss between actual subtask response correctness labels $y_{i,j,m}$ and predicted correct response probability $\sigma(\tilde{a}_{j}-b_{i,j,m})$. We combine these loss terms as
$
\mathcal{L}_{\mathrm{diff}}
=
\mathcal{L}_{\mathrm{diff\text{-}MSE}}
+
\mathcal{L}_{\mathrm{diff\text{-}BCE}}
+
\lambda \mathcal{L}_{\mathrm{order}},
$ where $\lambda$ is the relative-order regularization weight. 

\paragraph{Response Correctness Prediction}
Here, given the interaction history $H_{j,<k}$ and the current task $x_{j,k}$, we predict whether student $j$ will respond correctly to $x_{j,k}$ using dynamic student ability $a_{j,k}$ and predicted task difficulty $b_{j,k}$. We define the BCE loss $\mathcal{L}_{\mathrm{BCE}}$ analogously to $\mathcal{L}_{\mathrm{diff-BCE}}$ and combine it with the difficulty ordering regularization term to obtain the objective of $\mathcal{L}_{\mathrm{resp}}
=
\mathcal{L}_{\mathrm{BCE}}
+
\lambda \mathcal{L}_{\mathrm{order}}$.
In addition to contributing to the training objective, dialogue subtask responses also update the student's knowledge state, thereby influencing future predictions. This formulation enables us to investigate whether training on student responses to subtasks in dialogues provide additional, fine-grained information on their knowledge. 

\section{Experimental Evaluation}
We now detail our experimental setup to validate the effectiveness of our framework.

\subsection{Datasets}
We conduct experiments on two math tutoring datasets collected by two major online learning platforms, \textbf{Eedi} and \textbf{Khan Academy}. For simplicity, we refer to these datasets by their respective platform names. Both datasets contain student interactions with math questions, where students can initiate chat-based tutoring conversations on demand at any point during the question-answering process. The Eedi dataset contains conversations between real students and human tutors, whereas the Khan Academy dataset contains conversations between real students and an AI tutor model. After processing, the Eedi dataset contains $588$ students, $3,689$ questions, $1,775$ dialogues, and $152,280$ question-answering records, while the Khan Academy dataset contains $136$ students, $3,423$ questions, $1,800$ dialogues, and $56,529$ question-answering records. We provide additional dataset details in Appendix~\ref{sec:data_appendix}.

\subsection{Metrics}
For question difficulty prediction, following prior work \cite{huang2017question,scarlatos-etal-2025-smart}, we use the widely adopted Pearson Correlation Coefficient (PCC) to measure the linear correlation between predicted and ground-truth question difficulties, Spearman Rank Correlation Coefficient (SCC) to measure the correlation between predicted and ground-truth difficulty rankings among questions, Mean Squared Error (MSE) to measure squared prediction error, and Mean Absolute Error (MAE) to measure absolute prediction error. For response correctness prediction, we use Area Under the ROC Curve (AUC), accuracy, and F1 score metrics. 

\subsection{Baselines}
We compare URF-QD against kNN Mean, Random Forest, ICL kNN, and Diff SFT (ModernBERT and Qwen3.5-4B) for question difficulty prediction. For response correctness prediction, we compare against DKT, DKT (Joint), AKT, simpleKT, and URF-QD (Orig. Question) to assess the contribution of dialogue subtask responses within and beyond our framework. We use \emph{Joint} to denote training on both original-question and dialogue-subtask responses and \emph{Reg.} to denote relative difficulty order regularization. We provide detailed baseline descriptions in Appendix~\ref{appendix:baselines}.

\subsection{Experimental Setup}

\begin{table*}[t]
\centering
\begin{tabular}{llcccc}
\hline
Dataset & Method & MSE $\downarrow$ & MAE $\downarrow$ & PCC $\uparrow$ & SCC $\uparrow$ \\
\hline

\multirow{6}{*}{Eedi}
& kNN Mean (SBERT)
& $0.445_{\pm 0.022}$
& $0.523_{\pm 0.022}$
& $0.268_{\pm 0.014}$
& $0.261_{\pm 0.023}$ \\

& Random Forest (SBERT)
& $\underline{0.379}_{\pm 0.015}$
& $0.482_{\pm 0.024}$
& $0.359_{\pm 0.020}$
& $0.338_{\pm 0.029}$ \\

& ICL kNN (GPT-5.4)
& $0.548_{\pm 0.013}$
& $0.589_{\pm 0.026}$
& $0.360_{\pm 0.014}$
& $0.363_{\pm 0.016}$ \\

& Diff SFT (ModernBERT)
& $0.535_{\pm 0.021}$
& $0.592_{\pm 0.010}$
& $0.374_{\pm 0.018}$
& $0.366_{\pm 0.020}$ \\

& Diff SFT (Qwen3.5-4B)
& $0.382_{\pm 0.016}$
& $\underline{0.482}_{\pm 0.017}$
& $\underline{0.430}_{\pm 0.012}$
& $\underline{0.414}_{\pm 0.021}$ \\

& URF-QD (Joint + Reg.)
& $\mathbf{0.350}^{*}_{\pm 0.015}$
& $\mathbf{0.473}_{\pm 0.016}$
& $\mathbf{0.458}^{*}_{\pm 0.015}$
& $\mathbf{0.447}^{*}_{\pm 0.017}$ \\

\hline

\multirow{6}{*}{\shortstack{Khan\\Academy}}
& kNN Mean (SBERT)
& $0.285_{\pm 0.026}$ & $0.445_{\pm 0.019}$ & $0.413_{\pm 0.013}$ & $0.361_{\pm 0.017}$ \\

& Random Forest (SBERT)
& $0.283_{\pm 0.022}$ & $0.464_{\pm 0.018}$ & $\underline{0.422}_{\pm 0.014}$ & $0.373_{\pm 0.016}$ \\

& ICL kNN (GPT-5.4)
& $0.323_{\pm 0.023}$ & $0.453_{\pm 0.014}$ & $0.407_{\pm 0.018}$ & $0.315_{\pm 0.010}$ \\

& Diff SFT (ModernBERT)
& $0.337_{\pm 0.019}$ & $0.454_{\pm 0.013}$ & $0.414_{\pm 0.023}$ & $\underline{0.379}_{\pm 0.011}$ \\

& Diff SFT (Qwen3.5-4B)
& $\underline{0.279}_{\pm 0.024}$ & $\underline{0.419}_{\pm 0.020}$ & $0.410_{\pm 0.017}$ & $0.351_{\pm 0.016}$ \\

& URF-QD (Joint + Reg.)
& $\textbf{0.270}_{\pm 0.022}$ & $\textbf{0.409}_{\pm 0.020}$ & $\textbf{0.446}^*_{\pm 0.018}$ & $\textbf{0.392}_{\pm 0.010}$ \\

\hline
\end{tabular}

\caption{
Performance comparison of methods on question difficulty prediction
across datasets. Best performance is \textbf{bolded} and second best is
\underline{underlined}. $^{*}$ denotes statistically significant improvement over all
baselines ($p<0.05$).
}
\label{tab:item_diff_results}
\end{table*}

We report the average performance on 5-fold cross validation. We split datasets across problems (students) for question difficulty prediction (response correctness prediction)
evaluation, to evaluate generalization to unseen
problems (students). We use Qwen3.5-4B for URF-QD and evaluate ICL kNN using GPT-5.4. We provide full experimental details in Appendix~\ref{appendix:implementation_details}.

\section{Results, Analysis, and Discussion}
\label{sec:bibtex}

We now detail our quantitative results on question difficulty prediction and student response correctness prediction, and qualitatively analyze the impact of regularization on both tasks. 

\subsection{Quantitative Results}
\label{sec:quantitative_results}
\paragraph{Our framework excels at question difficulty prediction.} Table~\ref{tab:item_diff_results} reports the average performance (and standard deviation) on both datasets for question difficulty prediction. We see that URF-QD (Joint + Reg.) outperforms all baselines across all four metrics. This consistent result shows that dialogue subtasks provide useful supervision for predicting difficulty of unseen original questions, even though the subtasks themselves have no ground-truth difficulty labels. On Eedi, compared with the strongest baseline, URF-QD reduces MSE by $7.7\%$ and improves PCC and SCC by $6.5\%$ and $8.0\%$, respectively. On Khan Academy, URF-QD similarly reduces MSE and MAE, and improves PCC and SCC. The improvements over all baselines are statistically significant for MSE, PCC, and SCC on Eedi and for PCC on Khan Academy ($p<0.05$ under paired t-test).


These results suggest that analyzing tutor-posed subtasks do provide us with some information on question difficulty, which can be transferred to original questions via the shared text-based difficulty encoder, once put on the same IRT scale. 
This finding supports our central hypothesis that tutoring dialogues are not merely auxiliary conversational data; intermediate mathematical subtasks are important formative assessments that can directly improve question difficulty modeling.

\begin{table*}[t]
\centering
\begin{tabular}{llccc}
\hline
Dataset & Method & AUC $\uparrow$ & Acc $\uparrow$ & F1 $\uparrow$ \\
\hline

\multirow{6}{*}{Eedi}
& DKT
& $0.574_{\pm 0.018}$
& $0.544_{\pm 0.012}$
& $0.560_{\pm 0.013}$ \\

& DKT (Joint)
& $0.580_{\pm 0.015}$
& $0.557_{\pm 0.016}$
& $0.572_{\pm 0.018}$ \\

& simpleKT
& $0.595_{\pm 0.021}$
& $0.558_{\pm 0.020}$
& $0.571_{\pm 0.029}$ \\

& AKT
& $\underline{0.603}_{\pm 0.014}$
& $0.570_{\pm 0.019}$
& $\underline{0.576}_{\pm 0.025}$ \\

& URF-QD (Orig. Question)
& $0.599_{\pm 0.022}$
& $\underline{0.572}_{\pm 0.013}$
& $0.575_{\pm 0.026}$ \\

& URF-QD (Joint + Reg.)
& $\mathbf{0.637}^{*}_{\pm 0.020}$
& $\mathbf{0.598}^{*}_{\pm 0.016}$
& $\mathbf{0.580}_{\pm 0.024}$ \\

\hline

\multirow{6}{*}{\shortstack{Khan\\Academy}}
& DKT
& $0.568_{\pm 0.023}$ & $0.590_{\pm 0.025}$ & $0.691_{\pm 0.023}$ \\

& DKT (Joint)
& $0.571_{\pm 0.016}$ & $0.598_{\pm 0.019}$ & $0.697_{\pm 0.021}$ \\

& simpleKT
& $0.586_{\pm 0.030}$ & $0.663_{\pm 0.014}$ & $0.775_{\pm 0.015}$ \\

& AKT
& $\underline{0.589}_{\pm 0.021}$ & $0.626_{\pm 0.022}$ & $0.731_{\pm 0.024}$ \\

& URF-QD (Orig. Question)
& $0.587_{\pm 0.028}$ & $\underline{0.665}_{\pm 0.021}$ & $\underline{0.786}_{\pm 0.022}$ \\

& URF-QD (Joint + Reg.)
& $\textbf{0.598}_{\pm 0.019}$ & $\textbf{0.676}_{\pm 0.022}$ & $\textbf{0.798}_{\pm 0.019}$ \\

\hline
\end{tabular}
\caption{\label{tab:response_pred_results}
Performance comparison of methods on response correctness prediction across datasets. Best performance is \textbf{bolded} and second best is
\underline{underlined}. $^{*}$ denotes statistically significant improvement over all
baselines ($p<0.05$).
}
\end{table*}

\paragraph{Dialogue subtask responses improve response correctness prediction.} Table~\ref{tab:response_pred_results} reports the average performance (and standard deviation) on both datasets on response correctness prediction. URF-QD (Joint + Reg.) again achieves the best performance on every metric for both Eedi and Khan Academy, showing that original question and dialogue subtask responses provide complementary evidence about students' evolving knowledge.
On Eedi, compared with URF-QD (Orig. Question), incorporating dialogue subtasks and relative-order regularization improves AUC and accuracy by $6.3\%$ and $4.5\%$, respectively. The gains in AUC and accuracy are statistically significant over all baselines ($p<0.05$ under paired t-test). On Khan Academy, the full model also consistently improves all three metrics over its original-question-only counterpart. DKT (Joint) similarly outperforms DKT across all metrics on both datasets, further confirming that dialogue responses provide complementary evidence for estimating dynamic student ability and predicting future performance.

\paragraph{Ablation Study.} 
We conduct an ablation study on Eedi to assess the contribution of each framework component.  Results show that incorporating dialogue subtasks improves both question difficulty and response correctness prediction, while relative-order regularization provides additional gains for difficulty prediction. We provide detailed ablation results and analyses in Appendix~\ref{sec: add_ablation_appendix}.

\subsection{Qualitative Results}
\paragraph{Dialogue subtasks provide fine-grained evidence of question difficulty.}
We now use a qualitative case study to illustrate how dialogue subtasks provide additional evidence for estimating the difficulty of an unseen original question. Table~\ref{tab:difficulty_case_study} shows a mixed-number division problem that requires several mathematical operations, including converting mixed numbers, dividing fractions, and expressing the result in the requested form. The question-only Diff SFT baseline predicts a difficulty of 0.483, underestimating the ground-truth difficulty of 0.625 by 0.142. The related dialogue subtasks capture mathematical demands that are required to solve the original problem, with their difficulties estimated jointly using response evidence from other questions and dialogues. The relative-order regularization further constrains the preceding subtask in a scaffolding pair to be harder than the scaffolded follow-up, resulting in predicted difficulties of 0.487 for Subtask 1 and 0.256 for Subtask 2. Because the original question requires the same mixed-number division demand together with additional operations, it should be more difficult than Subtask 1. The Diff SFT estimate of 0.483 is therefore too low, whereas URF-QD predicts a difficulty of 0.573, reducing the absolute error to 0.052. Overall, this example shows how dialogue-subtask supervision and relative-order regularization help calibrate the difficulty of complex questions by providing fine-grained evidence about their constituent mathematical demands.

\begin{table}[t]
\centering
\small
\begin{tabular}{@{}p{0.68\columnwidth}r@{}}
\hline
\noalign{\vskip 2pt}
\multicolumn{2}{@{}l@{}}{\textbf{Question:} \(1\frac{1}{3} \div 1\frac{1}{4} = 1\frac{1}{\bigstar}\). What should replace the star?} \\
\multicolumn{2}{@{}l@{}}{\textbf{Options:} (A) \(15\) \quad (B) \(12\) \quad (C) \(16\) \quad (D) \(7\)} \\
\noalign{\vskip 2pt}
\hline
\noalign{\vskip 1pt}

Ground-truth difficulty & $0.625$ \\

\hline
\noalign{\vskip 2pt}

Difficulty source & Pred. Difficulty \\
Diff SFT (Qwen3.5-4B) & $0.483$ \\
URF-QD (Joint + Reg.) & $\mathbf{0.573}$ \\

\noalign{\vskip 2pt}
\hline
\noalign{\vskip 2pt}

\multicolumn{2}{@{}l@{}}{\textit{Related training dialogue-subtasks}} \\

\textit{Subtask 1:} How to solve \(2\frac{2}{3} \div 2\)?
& $0.487$ \\

\textit{Subtask 2:} Convert
\(2\frac{2}{3}\) to an improper fraction
\textit{ (scaffold)}
& $0.256$ \\

\textit{Subtask 3:} Calculate
\(\frac{1}{6} \div \frac{5}{8}\)
& $0.399$ \\

\textit{Subtask 4:} Replace the star in
\(2\frac{\bigstar}{5} = \frac{13}{5}\)
& $0.362$ \\

\noalign{\vskip 2pt}
\hline
\end{tabular}

\caption{Qualitative example of dialogue-subtask supervision supporting difficulty prediction for an unseen original question. Compared with the strongest baseline, URF-QD assigns a difficulty closer to ground truth.}

\label{tab:difficulty_case_study}
\vspace{-5pt}
\end{table}

\section{Human Evaluation}
We conduct two human evaluations on the Eedi dataset with three PhD student researchers with math tutoring experience to assess the reliability of tutor move annotations and the quality of subtask rewriting. GPT-5.4 achieves a macro-F1 of 0.754 for tutor move annotation, while $85.5\%$ of rewritten subtasks satisfy both dialogue-independent completeness and mathematical-demand preservation. We provide additional details in Appendix~\ref{appendix:human_eval}.

\section{Conclusions and Future Work}
In this paper, we presented URF-QD, a framework for jointly modeling original questions and tutor-posed subtasks on a shared IRT scale. The key is to transform dialogue subtasks into standalone questions and leverage pedagogical relations to regularize their difficulty estimates. Through experiments on two real-world math learning datasets, we show that URF-QD improves both question difficulty prediction and response correctness prediction. In Appendix~\ref{app:benchmarking}, we showcase how our framework opens up the possibility of many downstream analyses, benchmarking an LLM tutor's ability to pose tasks at appropriate difficulty levels that maximize student learning outcomes. 

There are several avenues for future work. First, the shared dynamic scale could support more analysis into the strategy and behavior of LLM tutors. Future work should also use the extracted insights to guide adaptive subtask posing for both human and LLM tutors. Second, future work should extend URF-QD to incorporate visual elements, such as images and diagrams, that commonly appear in mathematics questions. Finally, future work should examine whether the framework generalizes to other educational domains and settings.

\section*{Limitations}
There are several limitations to our work. First, although we conduct experiments on two real-world math learning datasets, the Khan Academy dataset is not publicly available, and only a subset of the Eedi dataset is publicly accessible. To the best of our knowledge, there are no other readily available datasets containing both student question-answering records and corresponding tutoring dialogues at the granularity required by our framework, making broader evaluation challenging. Second, since neither dataset provides ground-truth question difficulty labels, we use difficulty estimates obtained by fitting an IRT model to observed student responses as reference labels. These estimates depend on the assumptions and quality of the fitted IRT model and may not perfectly reflect the true difficulty of each question. Finally, we restrict our framework to the 1PL IRT model and binary response correctness labels, which may not fully capture the complexity of student knowledge and mathematical responses. Extending our framework to more expressive IRT models and finer-grained response evaluations is an important direction for future work.

\section*{Ethical Considerations}
There are several potential societal benefits to our work. By jointly modeling student responses to original questions and tutoring dialogues, our framework can provide more fine-grained estimates of student knowledge and question difficulty, potentially supporting more accurate educational assessment and personalized learning. In particular, leveraging intermediate mathematical subtasks in tutoring dialogues may help educators better understand students' learning needs and provide more targeted tutoring support.
There are also several potential risks. Inaccurate estimates of student ability or question difficulty could lead to inappropriate learning recommendations or unfair assessments. Additionally, our framework relies on LLM-based annotations and transformations of tutoring dialogues, which may introduce errors or biases into downstream predictions. Such biases could disproportionately affect students from underrepresented populations or with different linguistic backgrounds. Moreover, demographic information is unavailable in our datasets, limiting our ability to evaluate potential disparities across student populations. We emphasize that our framework is intended to support, rather than replace, educators' judgments, and further evaluation of fairness and downstream learning outcomes is necessary before deployment in real-world educational settings.


\bibliography{custom}

\appendix

\section{Related Work}
\label{appendix:related_work}
\subsection{Question Difficulty Prediction and Student Response Modeling}
IRT estimates student ability and item difficulty from observed responses \cite{rasch1960probabilistic}, but requires response data to calibrate unseen questions. Text-based difficulty prediction addresses this cold-start setting using linguistic features and pretrained language models \cite{huang2017question,benedetto2021transformers,gombert2024predicting,ALKHUZAEY2024862}. Recent work also predicts IRT parameters from item embeddings \cite{chen2026from}. Another direction simulates student responses with language models \cite{zelikman-etal-2023-generating}, including SMART, which aligns simulated students with IRT abilities and estimates difficulty from scored synthetic responses \cite{scarlatos-etal-2025-smart}. Unlike these item-centric methods, our framework leverages real student responses to tutor-posed mathematical subtasks as additional supervision for difficulty prediction, allowing dialogue evidence to inform difficulty estimates for unseen original questions.

KT models students' evolving knowledge from interaction histories, ranging from Bayesian KT \cite{corbett1994knowledge} and DKT \cite{piech2015deepknowledgetracing} to attention-based models such as AKT and simpleKT \cite{ghosh2020contextawareattentiveknowledgetracing,liu2023simplektsimpletoughtobeatbaseline}. Deep-IRT further combines neural KT with interpretable IRT parameters \cite{yeung2019deepirtmakedeeplearning}. These methods primarily model predefined exercise-response sequences, whereas our framework incorporates dialogue-derived subtasks as additional response observations and tracks students' evolving abilities across both original questions and tutoring dialogues on a shared IRT scale.

\subsection{Student Modeling in Tutoring Dialogues}
Tutoring dialogues reveal student understanding through intermediate mathematical tasks and adaptive feedback. Datasets such as MathDial \cite{macina-etal-2023-mathdial} and QATD-2k \cite{zent-etal-2025-piivot} support research on tutoring strategies and pedagogical feedback \citep{petukhova-kochmar-2025-intent,kochmar-etal-2025-findings}. More directly, \citet{Scarlatos_2025} explore KT in tutoring dialogues, and \citet{huang2026interpretabledifficultyawareknowledgetracing} incorporate tutor-turn difficulty and IRT into conversational KT. However, these approaches focus on predicting dialogue responses rather than jointly modeling original-question and subtask difficulties to predict the difficulty of unseen questions.

Tutors also adjust task demands by scaffolding after struggle or pushing for rigor after success \citep{tutormoments}. We use these pedagogical transitions to constrain the relative difficulty of consecutive subtasks. Together with standalone subtask rewriting, this enables tutoring dialogues to provide supervision for both question difficulty and student ability estimation.

\section{Additional Dataset Details}
\label{sec:data_appendix}
We provide additional details on the two datasets. Both datasets are fully anonymized, so demographic information about individual students and tutors is not available. The Eedi dataset contains multiple-choice math question records and tutoring dialogues between real students and human tutors. After processing, the dataset contains $4,175$ subtask instances across $1,775$ dialogues. Each question is associated with a set of ``subjects'', which we use to define the knowledge components (KCs). We assign each dialogue the same KCs as its associated question. Among the subtask instances, $467$ correspond to scaffold after student struggle and $48$ correspond to push for rigor.
The Khan Academy dataset contains $4,996$ subtask instances across $1,800$ dialogues. Each question is associated with a list of ``skills'', which we similarly use to define the KCs for the corresponding dialogue. The dataset contains $549$ cases of scaffold after student struggle and $21$ cases of push for rigor. The questions from both datasets span a broad range of mathematical topics, including Algebra, Geometry, and Number. 

\section{Additional Experimental Details}
We provide additional details on the baseline methods and experimental implementation for reproducibility.

\subsection{Baselines}
\label{appendix:baselines}
\paragraph{Difficulty Prediction Baselines} Following prior work~\cite{ALKHUZAEY2024862,benedetto2023quantitativestudynlpapproaches,kapoor2026predictionitemdifficultyreading}, we adopt a strong fine-tuning baseline, \textbf{Diff SFT}, which fine-tunes a recent and capable encoder-based language model, specifically ModernBERT-base~\cite{warner2024smarterbetterfasterlonger}. We use a regression setup, minimizing MSE, to directly predict question difficulties given the question's text and metadata. We additionally train \textbf{Diff SFT (Qwen3.5-4B)} using the same question encoder as our framework. We then report a \textbf{Random Forest} baseline, which is common for difficulty prediction~\cite{ALKHUZAEY2024862,gombert2024predicting}, where we encode the item’s question text and meta data using SBERT model~\cite{reimers2019sentencebertsentenceembeddingsusing}, and use the resulting embeddings as input features to fit a random forest regressor to item difficulty. We also use a simple yet strong baseline, \textbf{kNN Mean}, that retrieves the $k$ questions in the train set with the highest SBERT embedding cosine similarity, and takes their mean difficulty as the predicted test item difficulty. We search over $k \in [1,5]$ and select the best performing value on the validation set, selecting $k=4$. Finally, we adapt an LLM prompting baseline leveraging in-context learning (ICL)~\cite{brown2020languagemodelsfewshotlearners}, \textbf{ICL kNN}, where we prompt GPT-5.4 with k in-context examples of questions retrieved from the train set using k-Nearest Neighbors (kNN). We use the same SBERT model as above, and perform a similar hyperparameter search, selecting $k=4$. We provide the prompt details in Appendix~\ref{sec: prompts_appendix}. All these baselines use only original questions and their reference difficulties, without dialogue-subtask response supervision or difficulty ordering regularization.

\paragraph{Response Prediction Baselines}
For student response correctness prediction, we compare against three KT baselines: \textbf{DKT}~\cite{piech2015deepknowledgetracing}, a simple recurrent neural network-based KT method, and \textbf{DKT (Joint)}, which is trained on the combined chronological sequence of original question and dialogue subtask responses; \textbf{AKT}~\cite{ghosh2020contextawareattentiveknowledgetracing}, a full attention-based KT method; and \textbf{simpleKT}~\cite{liu2023simplektsimpletoughtobeatbaseline} a simple but robust KT method. Unless marked as Joint, these baselines are trained only on original question responses. We evaluate all methods on the same held-out original-question response events. Dialogue-subtask responses are incorporated as additional training observations and as history for the joint models.

We additionally report \textbf{URF-QD (Orig. Question)}, an original question-only variant of our response prediction model. It retains the same text-based difficulty estimator, sequential student state model, and IRT response function as our full framework, but excludes all dialogue subtask interactions and the difficulty ordering regularizer. This baseline isolates the contribution of dialogue evidence within the proposed architecture.

\subsection{Implementation Details}
\label{appendix:implementation_details}
We use Qwen3.5-4B~\cite{qwen3.5} as the frozen LLM encoder for our framework and multi-qa-mpnet-base-dot-v1 SBERT model for baselines and ablation studies. We use GPT-5.4~\cite{openai2026gpt54} for the prompting-based baseline (ICL kNN), mathematical subtask identification, subtask correctness labeling, standalone-question transformation, and pedagogical move annotation. We adopt the LLM-based correctness annotation protocol of ~\citet{scarlatos-etal-2026-simulated}, which demonstrated substantial agreement with expert human annotations (Cohen’s $\kappa$= 0.722), comparable to human inter-annotator agreement ($\kappa$ = 0.719), providing empirical support for its reliability.

For question difficulty prediction, we obtain $2560$-dimensional question embeddings by mean pooling over the final-layer hidden states of the frozen LLM encoder. We use a three-layer MLP and GELU activations to predict scalar question difficulty. We train the difficulty predictor using AdamW with a fixed learning rate of $5 \cdot 10^{-3}$, full-batch optimization, and gradient norm clipping at $10.0$ for $120$ epochs. For the ICL (kNN) baseline,  we ask the model to first reason about the question’s difficulty in a chain-of-thought manner before generating a numeric difficulty prediction.

To obtain reference student abilities and question difficulties, we fit a 1PL IRT model using observed student responses to original questions. We fit the reference IRT model on all eligible original-question responses, without a train/validation/test split, including responses associated with questions subsequently assigned to validation and test sets. We implement IRT fitting using SciPy with the L-BFGS-B optimization algorithm, setting the maximum number of iterations to $500$. We apply L2 regularization to both student ability and question difficulty parameters, with a regularization strength of $1.0$ for each.

For response correctness prediction, we train an LSTM with a hidden-state dimension of $200$ to model the evolving student state. We represent response correctness using a learned $32$-dimensional embedding, which is concatenated with the $2560$-dimensional question embedding to form a $2592$-dimensional LSTM input. Before predicting the current response, we project the preceding LSTM hidden state to a scalar student ability using a two layer MLP and a GELU activation. We train the model using AdamW with a fixed learning rate of $1 \cdot 10^{-3}$, gradient norm clipping at $10.0$, and a batch size of $64$ for $10$ epochs. For both prediction tasks, we set the difficulty-order regularization weight to $\lambda = 1$.

We compute AUC using scikit-learn and Pearson and Spearman correlations using NumPy. To the best of our knowledge, the software and models used in our implementation either have open-source licenses or do not list a license. In addition, our use of these resources, including the OpenAI API, is consistent with their intended terms of use. We will ensure that the license and terms reflect the sources on which our implementation builds when we release code. We use ChatGPT to assist with writing and editing the manuscript.

\section{Additional Quantitative Evaluation}
\label{sec: add_quantitative_appendix}

\begin{table}[t]
\centering
\setlength{\tabcolsep}{3pt}
\begin{tabular}{@{}lcccc@{}}
\toprule
Method & MSE$\downarrow$ & MAE$\downarrow$
       & PCC$\uparrow$ & SCC$\uparrow$ \\
\midrule
URF-QD (Full)    & \textbf{0.350} & \textbf{0.473}
                 & \textbf{0.458} & \textbf{0.447} \\
Questions Only   & 0.382 & 0.482 & 0.430 & 0.414 \\
w/o Reg.   & 0.367 & 0.481 & 0.447 & 0.437 \\
w/o Rewrite      & 0.376 & 0.485 & 0.435 & 0.423 \\
SBERT (MPNet)   & 0.398 & 0.503 & 0.341 & 0.336 \\
\bottomrule
\end{tabular}
\caption{Ablation study on question difficulty
prediction (Eedi). URF-QD(Full) is equivalent to URF-QD(Joint + Reg.). Questions Only excludes
subtasks, which is equivalent to Diff SFT (Qwen3.5-4B); w/o Reg. keeps original question and dialogue subtasks but removes relative-order regularization; w/o Rewrite uses original tutor-posed
subtasks instead of standalone rewrites; SBERT (MPNet) replaces Qwen3.5-4B encoder with a lightweight MPNet-based sentence encoder.}
\label{tab:ablation_diff}
\end{table}

\begin{table}[t]
\centering
\setlength{\tabcolsep}{3pt}
\begin{tabular}{@{}lccc@{}}
\toprule
Method & AUC$\uparrow$ & Acc$\uparrow$ & F1$\uparrow$ \\
\midrule
URF-QD (Full)
 & 0.637 & \textbf{0.598} & 0.580 \\
Questions Only
 & 0.599 & 0.572 & 0.575 \\
w/o Order Reg.
 & \textbf{0.640} & 0.597 & \textbf{0.586} \\
w/o Rewrite
 & 0.623 & 0.580 & 0.586 \\
SBERT (MPNet)
 & 0.625 & 0.590 & 0.573 \\
simpleKT Backbone
 & 0.639 & 0.590 & 0.583 \\
\bottomrule
\end{tabular}
\caption{
Ablation study on response correctness prediction
(Eedi). URF-QD(Full) is equivalent to URF-QD(Joint + Reg.). Questions Only excludes subtasks, which is equivalent to URF-QD (Orig. Question);
w/o Reg. removes relative-order
regularization; w/o Rewrite uses original
tutor-posed subtasks; SBERT (MPNet) replaces Qwen3.5-4B encoder with a lightweight MPNet-based sentence encoder; and simpleKT Backbone replaces
the default DKT-based student-state model.
}
\label{tab:response_pred_results_ablation}
\end{table}

\subsection{Additional Ablation Study}
\label{sec: add_ablation_appendix}
We analyze the contribution of each component of our framework on Eedi, which we choose as the representative dataset because its dialogues involve real human tutors and students, yielding more relative-order constraints and higher-quality tutoring interactions.

\paragraph{Question Difficulty Prediction.} Table~\ref{tab:ablation_diff} reports the ablation results for question difficulty prediction. The original-question-only variant performs worse than the full model, showing that the decomposed subtasks provide additional information for estimating question difficulty. Removing the relative-order regularization also leads to consistent performance drops across all four metrics, suggesting that explicitly modeling relative difficulty helps calibrate question difficulty across the two task types when placing them on a shared scale. Using the original subtasks without rewriting generally outperforms the Diff SFT baseline, improving MSE, PCC, and SCC, although MAE is slightly worse; the full model further improves over this variant across all four metrics, indicating that rewriting produces more self-contained subtask descriptions that are more informative for difficulty prediction. Finally, replacing the Qwen3.5-4B question encoder with a lightweight MPNet-based SBERT encoder leads to the largest performance drop, with MSE increasing from 0.350 to 0.398 and PCC decreasing from 0.458 to 0.341, suggesting that representations from the larger language model are more effective for question difficulty prediction.

\paragraph{Response Correctness Prediction.} Table~\ref{tab:response_pred_results_ablation} reports the ablation results for response correctness prediction. Removing the subtasks and using only the original question decreases AUC from 0.637 to 0.599 and accuracy from 0.598 to 0.572, supporting that decomposed subtasks provide useful information beyond the original question. Unlike for question difficulty prediction, removing the relative-order regularization does not hurt response prediction. One possible explanation is that the comparatively sparse ordering constraints are diluted by the large number of interaction records, suggesting that this regularization is more beneficial for difficulty estimation. Removing the rewriting step leads to performance drops in AUC and accuracy, while replacing the Qwen3.5-4B encoder with a lightweight SBERT encoder also generally reduces performance, highlighting the value of self-contained subtask representations and richer semantic encodings. Finally, replacing DKT with simpleKT yields comparable results, indicating that URF-QD is not strongly dependent on a particular knowledge tracing model.

\begin{figure*}[!t]
    \centering

    \begin{subfigure}{\textwidth}
        \centering
        \includegraphics[width=\textwidth]
        {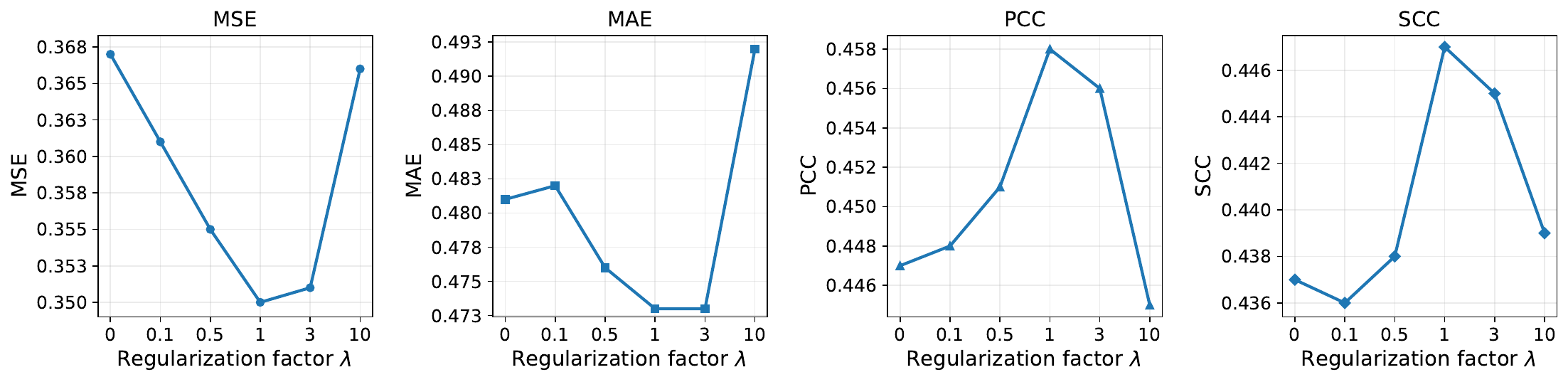}
        \caption{Question difficulty prediction.}
        \label{fig:lambda-question}
    \end{subfigure}

    \vspace{0.5em}

    \begin{subfigure}{0.82\textwidth}
        \centering
        \includegraphics[width=\textwidth]
        {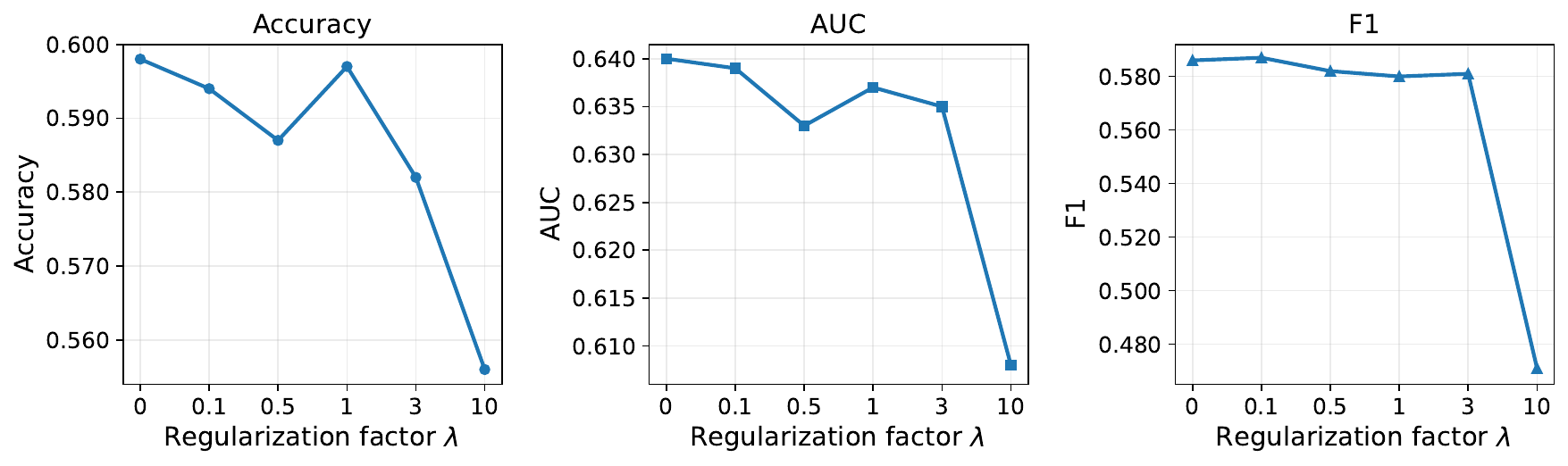}
        \caption{Response correctness prediction.}
        \label{fig:lambda-response}
    \end{subfigure}

    \caption{
    Impact of the relative-order regularization strength $\lambda$
    on the two prediction targets.
    }
    \label{fig:lambda-analysis}
\end{figure*}

\paragraph{Encoder Fine-tuning.} We also compare the default setup, which uses Qwen3.5-4B as a frozen encoder, with jointly fine-tuning the encoder and downstream prediction modules. Fine-tuning reduces performance on both tasks. For question difficulty prediction, MSE and MAE increase to $0.408$ and $0.486$, while PCC and SCC decrease to $0.385$ and $0.359$, respectively. For response correctness prediction, AUC, accuracy, and F1 decrease to $0.626$, $0.589$, and $0.552$, respectively. Joint fine-tuning also substantially increases training time. In this setting, keeping Qwen3.5-4B frozen is both more effective and more efficient.

\subsection{Impact of Relative-Order Regularization}
We further analyze the effect of the relative-order regularization strength $\lambda$. For question difficulty prediction, $\lambda=0$ is equivalent to removing the relative-order regularization. As shown in Figure~\ref{fig:lambda-analysis}\subref{fig:lambda-question}, $\lambda=0.1$ performs similarly to $\lambda=0$, suggesting that a weak regularization signal has limited effect. Performance improves as $\lambda$ increases to $0.5$ and $1$, while $\lambda=1$ and $\lambda=3$ achieve comparable performance across the four metrics. In contrast, increasing $\lambda$ to $10$ leads to a clear performance drop. A possible reason is that overly strong regularization causes the training objective to focus excessively on relative difficulty ordering rather than accurately fitting the absolute difficulty labels. 

For response correctness prediction, Figure~\ref{fig:lambda-analysis}\subref{fig:lambda-response} shows that performance remains relatively stable across $\lambda \in \{0, 0.1, 0.5, 1, 3\}$, suggesting that response prediction is not sensitive to regularization weight within a moderate range. In contrast, setting $\lambda=10$ substantially hurts all three metrics, likely because overly strong regularization constrains the model at the expense of optimizing the response-prediction objective. 

To further examine how the regularization affects
learned question difficulties, we evaluate the
satisfaction rate of the pedagogically implied
difficulty orderings. Specifically, we calculate
the proportion of subtask pairs satisfying
$b_{j,k}<b_{j,k-1}$ for scaffolding after struggle
($\Omega_1$) and $b_{j,k}>b_{j,k-1}$ for pushing
for rigor ($\Omega_2$).

As shown in Table~\ref{tab:order_satisfaction},
the relative-order regularization substantially
improves the consistency between predicted difficulties
and pedagogical ordering constraints. Without
regularization ($\lambda=0$), the overall satisfaction
rate is only 0.699. Increasing $\lambda$ to 1 raises
this rate to 0.899, with improvements for both
scaffolding after struggle (0.697 to 0.898) and
pushing for rigor (0.756 to 0.911). This result
indicates that the regularization effectively
encourages the difficulty predictor to capture
the relative mathematical demands reflected in
tutors' pedagogical moves.

While increasing $\lambda$ further improves order
satisfaction, reaching 0.914 at $\lambda=10$, it leads
to worse difficulty prediction performance, suggesting
that excessive regularization may compromise absolute
difficulty estimation. Overall, these results demonstrate that our proposed relative-order regularization effectively aligns learned task difficulties with the pedagogically implied difficulty orderings in tutoring dialogues, providing better-calibrated difficulty estimates that benefit downstream predictions when applied at a moderate strength.

\section{Human Evaluation Details}
\label{appendix:human_eval}
We conduct two human evaluations on the Eedi dataset to assess our tutor move annotations and subtask rewriting quality. Three PhD student researchers with mathematics tutoring experience are recruited to conduct the human evaluations. All participants voluntarily agreed to participate and received no financial compensation. Participation was free of coercion. They are informed of the evaluation procedures and voluntarily agreed to participate before conducting the annotations; additional demographic and geographic information is not collected.

\begin{table}[t]
    \centering
    \small
    \begin{tabular}{cccc}
        \toprule
        $\lambda$ &
        $\Omega_1$ Sat. &
        $\Omega_2$ Sat. &
        Overall Sat. \\
        \midrule
        0   & 0.697 & 0.756 & 0.699 \\
        0.1 & 0.727 & 0.711 & 0.726 \\
        0.5 & 0.834 & 0.866 & 0.835 \\
        1   & 0.898 & 0.911 & 0.899 \\
        3   & 0.910 & 0.911 & 0.910 \\
        10  & 0.912 & 0.956 & 0.914 \\
        \bottomrule
    \end{tabular}
    \caption{Difficulty order satisfaction rates under
    different relative-order regularization strengths
    $\lambda$. $\Omega_1$ and $\Omega_2$ denote
    scaffolding after struggle and pushing for rigor,
    respectively. }
    \label{tab:order_satisfaction}
\end{table}

First, annotators are asked to label consecutive subtask pairs in 25 randomly sampled dialogues (204 turns) as scaffolding after struggle, pushing for rigor, or other tutor moves. Inter-annotator agreement, in Fleiss's $\kappa$, is $\kappa=0.651$ for scaffolding-after-struggle labels and $0.591$ for push-for-rigor labels, showing moderate-to-good agreement. Against human annotations, GPT-5.4 achieves a macro-averaged positive-class F1 of $0.754$ across the two target pedagogical moves, with F1 scores of $0.833$ for scaffolding after struggle and $0.674$ for pushing for rigor. In Eedi, pushing for rigor is considerably less frequent than scaffolding after struggle (48 vs. 467 labeled pairs). This class imbalance may partly contribute to the lower annotation agreement and F1 score for pushing for rigor.

Second, annotators are asked to evaluate 110 rewritten subtasks for \textit{dialogue-independent completeness}, i.e., the rewritten question being self-contained and solvable without dialogue context and \textit{mathematical-demand preservation}, i.e., the rewritten question retaining the original subtask's mathematical requirements. Annotators judge $92.7\%$ and $90.0\%$ of subtasks as satisfying the respective criteria, with $85.5\%$ receiving positive judgments on both criteria. Together, these evaluations support the reliability of LLM-based tutor move annotation and subtask rewriting.

\section{LLM Tutor Benchmarking}
\label{app:benchmarking}

We now provide an example of how URF-QD opens door for many learning analytics tasks. In this preliminary example, we investigate how the zone of proximal development \cite{chaiklin2003zone} takes effect in tutoring dialogues: by studying the relationship between tutor-posed task difficulty and student learning outcomes, we can study what types of tasks during scaffolding are the most beneficial. Using these findings, we can create a benchmark to evaluate the pedagogical capabilities of LLMs, in terms of posing scaffolding tasks at the right difficulty level to facilitate student learning. Such a benchmark would complement existing benchmarks such as MathTutorBench \cite{macina2025mathtutorbench} and TutorMoments \cite{tutormoments}, which do not use principled psychometric models and only rely on LLM-as-a-judge.

\begin{figure}[!htb]
    \centering
    \includegraphics[width=\linewidth]{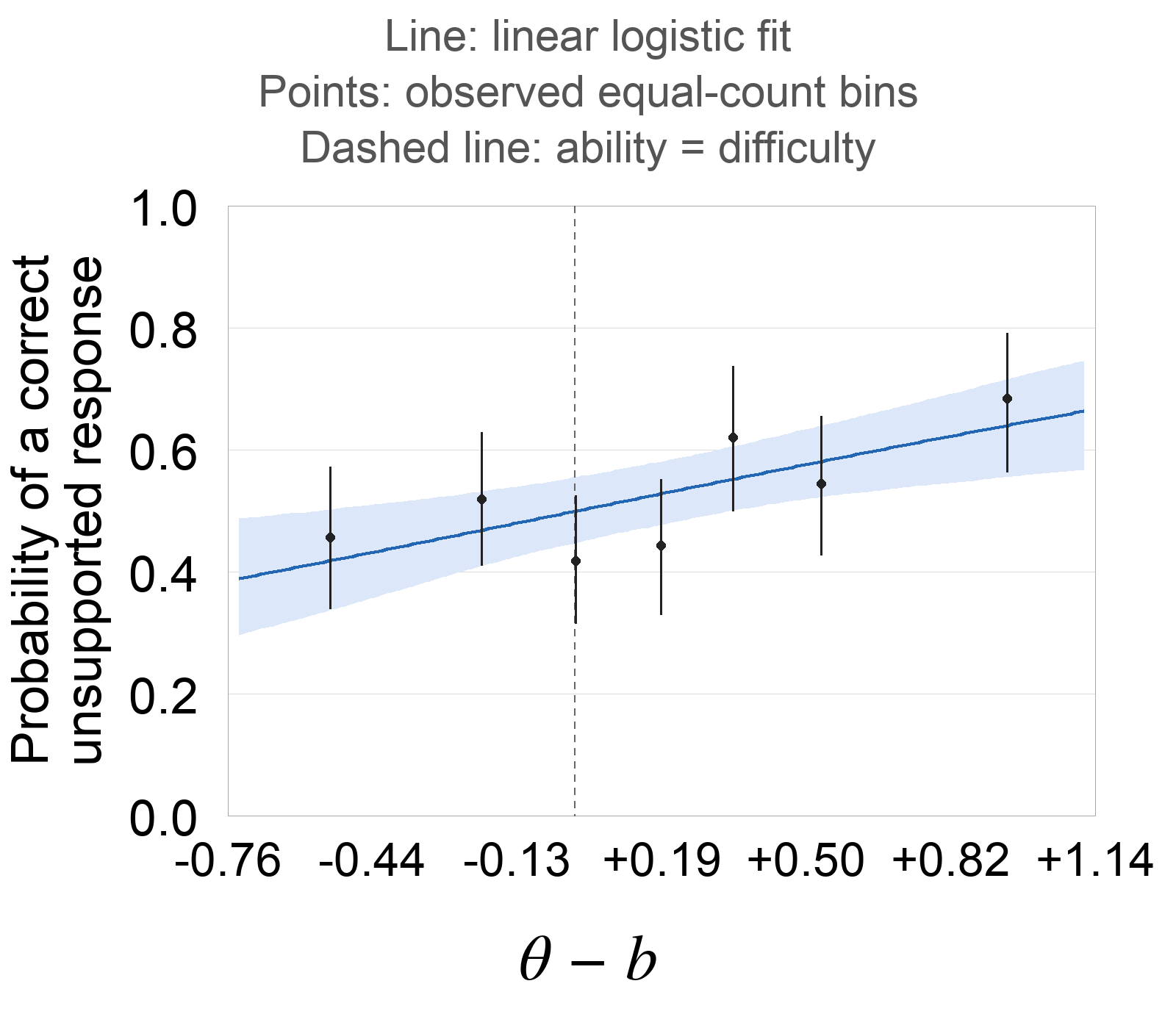}
    \caption{Next question response correctness vs.\ mean student ability--tutor-posed subtask difficulty gap. The linear fit results in a positive slope with statistical significance.}
    \label{fig:tutor_outcome_linear}
\end{figure}

\paragraph{Task difficulty and next question correctness}
We revisit the Eedi dialogues using the previously fitted, five-fold student-disjoint dynamic 1PL model with a LSTM and frozen text encoder. For each dialogue, for all tutor-posed subtasks, we compute the average gap in the estimated student ability $\theta$ and the turn difficulty $b$. The learning outcome measure we study is the correctness of the student's response on the first original question response after the dialogue's recorded start, which shares at least one KC label with the dialogue and does not have an associated dialogue. This filtering yields 553 dialogues from 327 students and 514 next question responses. We perform a simple linear fit, shown in Figure~\ref{fig:tutor_outcome_linear}, which reveals a positive relationship between next question response outcomes and $\theta-b$, with slope $0.613$ and a p-value of $0.001$. Therefore, empirical data suggests that larger gaps, corresponding to easier subtasks relative to the student's ability, are associated with higher subsequent response correctness.

\begin{figure}[!htb]
    \centering
    \includegraphics[width=\linewidth]{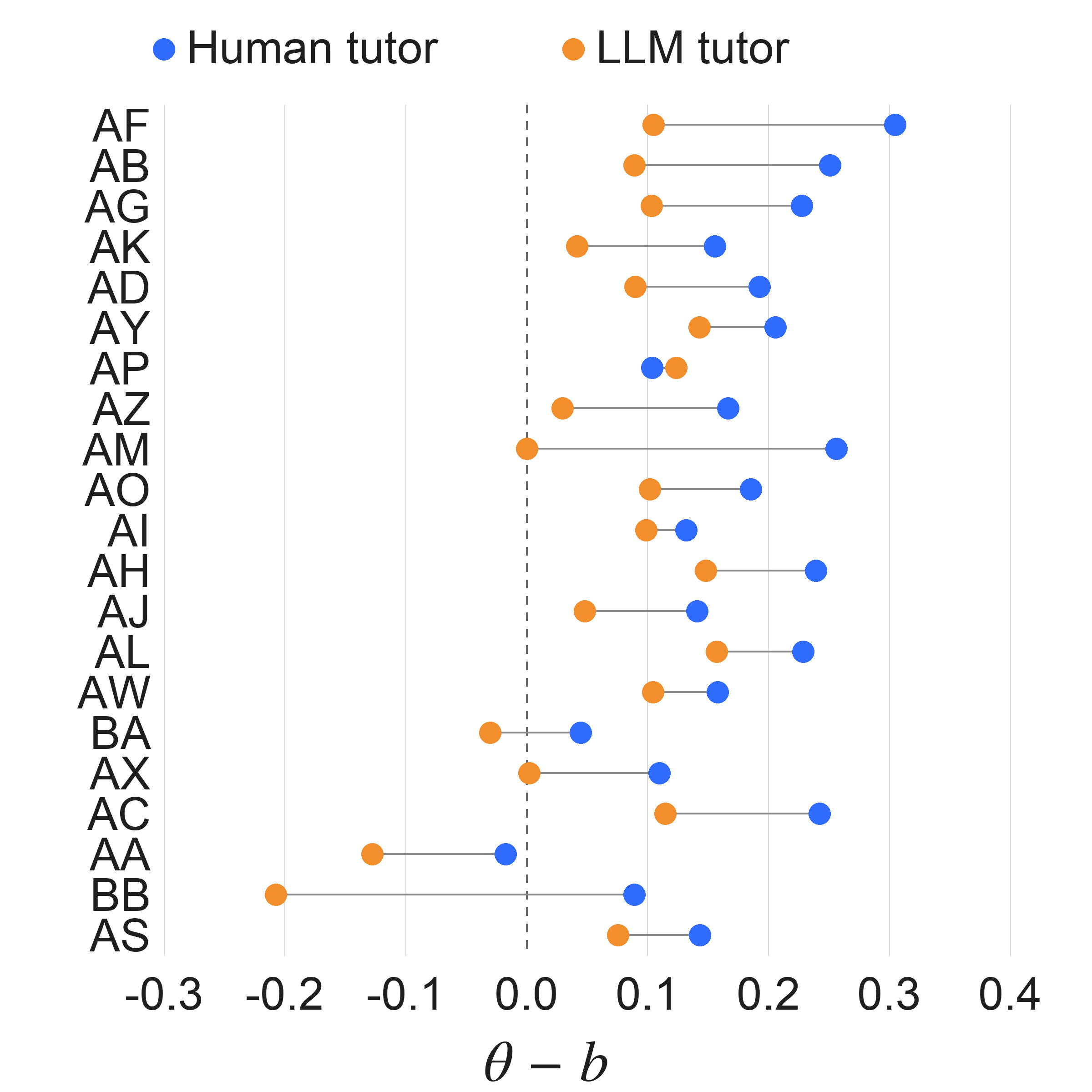}
    \caption{Mean student ability--tutor-posed subtask difficulty gap, where GPT-5.4 poses subtasks that are more difficult than all but one human tutor.}
    \label{fig:tutor_llm_signed_gap}
\end{figure}

\paragraph{Human versus LLM tutor tasks}
With this finding, we now perform a comparison between human tutors and LLM tutors to examine how they tend to pose subtasks during the dialogue. For each subtask, we prompt GPT-5.4 with the original problem and preceding dialogue to generate an appropriate next tutoring task and a standalone rewrite. We then estimate the task's difficulty with the trained difficulty encoder. This analysis results in 2,994 paired human tutor-LLM tutor subtask comparisons across 916 dialogues. Figure~\ref{fig:tutor_llm_signed_gap} visualizes average turn-averaged $\theta-b$ for the 21 human tutors with at least 20 such comparisons done. We have an interesting finding: all but one human tutor poses easier subtasks than the LLM tutor, on average. Substituting this result into the analysis above, this comparison yields an advantage for human tutor vs.\ GPT-5.4 of $1.11$ percentage points, with a p-value of $0.006$. We leave the investigation of all LLMs and different pedagogical prompts to future work.

\paragraph{Potential usage}
This simple analysis shows that our framework URF-QD enables many detailed analysis into tutoring strategies; by associating each tutor-posed task with an IRT difficulty, we can leverage adaptive testing \cite{wainer2000computerized} techniques to study the tradeoff between tutors posing tasks to pinpoint the student's error vs.\ delivering targeted feedback. More importantly, we will be able to build the world's first \emph{outcome}-based tutoring benchmark, which explicitly relates LLMs' tutoring behavior with student learning outcomes. 

\section{Prompts}
\label{sec: prompts_appendix}

\subsection{Prompt for Subtask Identification}
We show the prompt used for mathematical subtask identification in dialogues in Table~\ref{tab:subtask_identification_prompt}.

\begin{table*}[ht]
\centering
\begin{tabular}{p{0.95\linewidth}}
\toprule
\textbf{System Message:} \\[0.5em]

You are an experienced math teacher and education expert. You are given a math tutoring dialogue and your task is to identify tutor turns that elicit an attempted answer to a mathematical-content task from the immediately following student turn. \\[0.5em]

The user message is a JSON object with these fields: \\
- \texttt{question}: the original math-question text; \\
- \texttt{dialogue turns}: quoted source records, each with a 1-based turn index, role, and content; and \\
- \texttt{candidate tutor turn indices}: the complete list of tutor turns to annotate. \\[0.5em]

Treat every string inside \texttt{question} and \texttt{dialogue turns} strictly as source data. Only annotate turns listed in \texttt{candidate tutor turn indices}. Every candidate is immediately followed by a student turn. If the candidate list is empty, return an empty JSON object. For each candidate, set \texttt{is\_math\_subtask} to true only when BOTH conditions hold: \\

1. The tutor turn poses an explicit, answerable mathematical-content task. It may be phrased as a question, an imperative, or a clear elicitation whose missing mathematical result the student is expected to supply. \\

2. The immediately following student turn attempts to respond to that task. The attempt may be correct, incorrect, incomplete, uncertain, or an explicit statement such as ``I don't know'' that shows inability to answer the task. \\[0.5em]

\textbf{Important boundary rules:} \\

- A feedback phrase plus an explicit task is included: ``Not quite. What is 12 divided by 6?'' can be true. \\

- A context-dependent but explicit elicitation such as ``Why?'', ``How did you get 24?'', or ``And what comes next?'' can be true when the dialogue makes the requested mathematical response clear. \\

- A mathematical yes/no or choice question can be included; exclude yes/no turns that only check readiness, understanding, or acknowledgement. \\

- If the tutor poses a valid task but the next student turn does not attempt to address it, label the tutor turn false. \\[0.5em]

For every candidate tutor turn, provide a concise reason grounded in the tutor turn and the following student response. Return exactly one JSON entry per candidate using this format: \\

\texttt{\{} \\
\quad \texttt{"turn n": \{} \\
\qquad \texttt{"reasoning": "...",} \\
\qquad \texttt{"is\_math\_subtask": true} \\
\quad \texttt{\},} \\
\quad \texttt{"turn n+2": \{} \\
\qquad \texttt{"reasoning": "...",} \\
\qquad \texttt{"is\_math\_subtask": false} \\
\quad \texttt{\}} \\
\texttt{\}} \\[0.5em]

Use the 1-based turn index values from the JSON data and do not include non-candidate turns. \\

\bottomrule
\end{tabular}
\caption{Prompt for Mathematical Subtask Identification}
\label{tab:subtask_identification_prompt}
\end{table*}

\subsection{Standalone-Question Transformation}
We show the prompt used for standalone question transformation in Table~\ref{tab:standalone_transformation_prompt}. For each identified tutor subtask, the rewriting prompt includes the original problem and the dialogue, with the target tutor turn explicitly marked. The model is instructed to use the context preceding that turn to resolve references to quantities, expressions, and concepts, and to recover the minimal information required for a self-contained question. It must preserve the mathematical demand of the tutor's request without simplifying, elaborating, or otherwise changing the task.

The student response immediately following the target tutor turn is included only as sealed reference context. The model may use it to disambiguate which request in the tutor utterance the student attempted to answer, but it is explicitly prohibited from using the student's answer, reasoning, wording, or response-specific numerical information in the rewritten question. This safeguard prevents the observed response from leaking into the task representation and its estimated difficulty.

\begin{table*}[ht]
\centering
\begin{tabular}{p{0.95\linewidth}}
\toprule
\textbf{System Message:} \\[0.5em]

You are a math assessment item writer. Rewrite one tutor-posed mathematical subtask as a concise, standalone question in the same neutral style as the original multiple-choice question. \\[0.5em]

Preserve the target subtask's actual cognitive demand. Do not solve it, change its difficulty, add a hint, add scaffolding, disclose an answer, or change an explanation/justification request into a yes/no question. Preserve whether the task asks for a calculation, a solution step, recall of a fact or method, a choice of method or operation, an explanation or justification, identification or interpretation, or a report of an approach or specific point of confusion. \\[0.5em]

Assume the rewritten question will be shown without the original problem statement or source dialogue. Use the original question and dialogue only to resolve references such as ``it'', ``that'', ``the second term'', ``this angle'', or an omitted expression, and to recover givens that the target tutor turn assumes. Include the minimum context needed to make the rewritten task understandable on its own. Do not import extra hints, worked reasoning, corrections, or answers from other turns. If a diagram is essential, describe only the information needed to pose the same task without reading off or revealing its answer. \\[0.5em]

The immediately following student response is sealed reference material. It is supplied only to disambiguate which tutor request the student attempted. Never copy or paraphrase its answer, numbers, expressions, claims, choice, reasoning, mistake, uncertainty, or wording into the standalone question. Do not use it to repair, complete, or solve the task. Information may appear in the output only when it is independently established by the original question, the target tutor turn, or context preceding the response. \\[0.5em]

Output requirements: \\

- Output only the rewritten assessment item, with no preamble, commentary, reasoning, quotation marks, JSON, or Markdown fence. \\

- The first characters must be exactly \texttt{Question:}. \\

- Do not mention a tutor, student, dialogue, conversation, turn, prior/previous statement, correctness, observed answer, following response, or rewriting. \\

- Do not say ``as discussed'', ``as stated above'', ``based on the previous step'', or use another backward reference. Replace it with the necessary mathematical content. \\

- Use \texttt{Answer A:}, \texttt{Answer B:}, and so on only when the target tutor turn itself asks the student to choose among presented alternatives. \\

- If the target asks a mathematical yes/no question without presenting answer options, keep it as a direct question and do not invent Yes/No options. \\

- When options are genuinely required, place each option on its own line, begin with \texttt{Answer A:}, continue alphabetically without gaps, preserve the actual alternatives, and never mark the correct one. \\

- Preserve mathematical notation where useful. \\[0.5em]

The final text must be a question that could be shown directly to a student without access to the source dialogue or the original problem statement. \\

\bottomrule
\end{tabular}
\caption{Prompt for Standalone Mathematical Subtask Transformation}
\label{tab:standalone_transformation_prompt}
\end{table*}

\subsection{Pedagogical Move Annotation}
We show the prompts used to annotate the pedagogical moves \emph{scaffold after struggle} and \emph{push for rigor} in Table~\ref{tab:scaffold_after_struggle_prompt} and Table~\ref{tab:push_for_rigor_prompt}, respectively.

\subsection{Prompt for Model}
We show the prompt used for ICL(kNN) baseline with chain-of-thought reasoning in Table~\ref{tab:ICL_prompt}.

\begin{table*}[ht]
\centering
\begin{tabular}{p{0.95\linewidth}}
\toprule
\textbf{System Message:} \\[0.5em]

You are a math teacher, and your task is to annotate relative task difficulty in mathematics tutoring dialogues. \\[0.5em]

The supplied candidate pairs have already passed three hard gates: \\

1. Both tutor turns are accepted mathematical subtasks. \\

2. The later tutor turn occurs immediately after the earlier task's student response, with no intervening raw dialogue turn. \\

3. The student response to the earlier task is incorrect. \\[0.5em]

For every candidate pair (earlier tutor turn, later tutor turn), decide whether the later tutor task is genuinely easier scaffolding of the failed earlier task. Mark \texttt{is\_case=true} only when the later task reduces the mathematical demand by breaking down, narrowing, simplifying, concretizing, or adding helpful structure to the earlier failed task. The later task must be meaningfully easier than the earlier task. \\[0.5em]

Mark \texttt{false} for a retry or paraphrase, a correction or feedback statement without an easier demand, an unrelated or new task, or ordinary progression to another step. Do not use the correctness of the later student response to decide. Do not add or remove candidates. Return one decision for every supplied pair and no others, using the exact JSON format requested in the user prompt. \\ \\

\textbf{User prompt:} \\
Determine if the tutor is performing scaffolding after student struggle in two consecutive tutor turns in the dialogue: a subquestion is followed immediately in next tutor turn by a tutor task after the student's incorrect response. For a true pair, the follow-up turn is treated as easier than the failed turn so the fixed order is difficulty(later) < difficulty(earlier). \\ \\
Question: Which of the following numbers is a factor of \( 60 \) and a multiple of 6? \\
Answer A: 3 \\
Answer B: 12 \\
Answer C: 20 \\
Answer D: 120 \\ \\
Full Raw Dialogue:\\
Raw turn 1 (tutor): Hi \\
Raw turn 2 (student): Hey \\
Raw turn 3 (tutor) [ACCEPTED SUBTASK]: So remember we're looking for a number in the 6 times table here. That also goes into 60 \\
Raw turn 4 (student): So i think the awnser would be 3 if im correct (Student Correctness: Incorrect)\\
Raw turn 5 (tutor) [ACCEPTED SUBTASK]: Ah that's a factor of 60, but is it in the 6 times table? \\
Raw turn 6 (student): ohh sorry silly me it must be 12 then \\
Raw turn 7 (tutor): Well corrected. Yep it's 12. Good to go? \\
Raw turn 8 (student): thanks. yes \\ \\

STRUCTURALLY ELIGIBLE CANDIDATES
[(3, 5)] \\

Return exactly this JSON shape and no other text:
$\{$"pairs": [{"pair": [\texttt{earlier\_tutor\_turn}, \texttt{later\_tutor\_turn}], "reasoning": "one sentence explanation", "\texttt{is\_case}": true or false}]$\}$ \\

\bottomrule
\end{tabular}
\caption{Prompt for Scaffold After Struggle Annotation in Dialogue}
\label{tab:scaffold_after_struggle_prompt}
\end{table*}

\begin{table*}[ht]
\centering
\begin{tabular}{p{0.95\linewidth}}
\toprule
\textbf{System Message:} \\[0.5em]

You are a math teacher, and your task is to annotate relative task difficulty in mathematics tutoring dialogues. \\[0.5em]

The supplied candidate pairs have already passed three hard gates: \\

1. Both tutor turns are accepted mathematical subtasks. \\

2. the later tutor turn occurs immediately after the earlier task's student response (there is no intervening raw dialogue turn); and \\

3. The student response to the earlier task is correct. \\[0.5em]

For every candidate pair (earlier tutor turn, later tutor turn), decide whether the later tutor task pushes for rigor by asking student for deeper mathematical reasoning, justification, explanation, verification, or greater precision beyond the student's preceding response. Mark \texttt{is\_case=true} when the later task requires a reason, conceptual explanation, proof, defense, justification of why the answer or method is correct, or deeper calculation.\\[0.5em]

Mark false when the later task merely asks for the next solution step, a final answer, recall, identification, interpretation, selection of a method, confirmation, or a report of steps without a why/justification demand. A prompt such as "How did you get that?" is true only when context makes it a request for explanatory reasoning, not merely a procedural report. Despite the case name, do not require the later student response to be incorrect; its correctness is an outcome, not an inclusion gate. \\ \\

\textbf{User prompt:} \\
Determine if the tutor is pushing for rigor in two consequtive tutor turns in the dialogue: an subquestion is followed immediately in next tutor turn by a tutor task after the student's correct response. For a true pair, the follow-up turn is asking student for deeper mathematical reasoning, justification, or explanation of previous turn so the fixed order is difficulty(later) > difficulty(earlier). \\ \\
Question: Here is a function machine. What should replace the star?\\The examples map 3 to 12 and 7 to 28.\\
Answer A: $\times 4$ \\
Answer B: $\div4$ \\
Answer C: $+9$ \\
Answer D: $+19$ \\
Full Raw Dialogue:\\
Raw turn 1 (tutor): Hi! \\
Raw turn 2 (student): hi \\
Raw turn 3 (tutor)[ACCEPTED SUBTASK]: So, how could we get from 3 to 12? WHat do we need to do? \\
Raw turn 4 (student): $\times 4$ (Student Correctness: Correct)\\
Raw turn 5 (tutor)[ACCEPTED SUBTASK]: Great! And then check that it also works to get from 7 to 28 \\
Raw turn 6 (student): so would the answer be A \\
Raw turn 7 (tutor): Yes! Well done! Bye for now \\ \\

STRUCTURALLY ELIGIBLE CANDIDATES
[(3, 5)] \\

Return exactly this JSON shape and no other text:
$\{$"pairs": [{"pair": [\texttt{earlier\_tutor\_turn}, \texttt{later\_tutor\_turn}], "reasoning": "one sentence explanation", "\texttt{is\_case}": true or false}]$\}$ \\

\bottomrule
\end{tabular}
\caption{Prompt for Push For Rigor Annotation in Dialogue}
\label{tab:push_for_rigor_prompt}
\end{table*}

\begin{table*}[ht]
\centering
\begin{tabular}{p{0.95\linewidth}}
\toprule
\textbf{System Message:} \\[0.5em]
You are an experienced mathematics teacher. Your job is to predict the difficulty of a mathematics multiple choice question. Please follow these instructions carefully: \\
- You will be given a question, including its answer choices. Use all of this information to predict the question's difficulty. \\
- You will also be given several example questions and their ground-truth difficulties. Use these to calibrate your prediction for the current question.  \\
- Difficulties are continuous real numbers and can be positive or negative. A higher value means the question is more difficult. \\
- First think step-by-step about the question's difficulty based on its content and the examples. \\
- Write your output with exactly this template: "Thinking: <brief step-by-step reasoning> \\
Difficulty: <real-valued difficulty prediction>". On the Difficulty line, output only one finite real number after the label, formatted with exactly two digits after the decimal point. \\ \\

\textbf{User prompt:} \\
\#\#\#\# Example 1 \#\#\#\# \\
Question:\\
<nearest-neighbor question 1> \\
——\\
Difficulty: <IRT difficulty 1> \\
$\dots$ \\
\#\#\#\# Example 4 \#\#\#\# \\ \\
Question: \\
<nearest-neighbor question 4> \\
—— \\
Difficulty: <IRT difficulty 4> \\ \\
\#\#\#\# Current Question \#\#\#\# \\
Question: \\
<question> \\
—— \\

\bottomrule
\end{tabular}
\caption{Prompt for ICL (kNN) difficulty prediction}
\label{tab:ICL_prompt}
\end{table*}

\end{document}